\documentclass[10pt, a4paper]{article}
\usepackage{lrec2022} 
\usepackage{multibib}
\newcites{languageresource}{Language Resources}
\usepackage{graphicx}
\usepackage{tabularx}
\usepackage{soul}
\usepackage{titlesec}
\titleformat{\section}{\normalfont\large\bfseries\center}{\thesection.}{1em}{}
\titleformat{\subsection}{\normalfont\SmallTitleFont\bfseries\raggedright}{\thesubsection.}{1em}{}
\titleformat{\subsubsection}{\normalfont\normalsize\bfseries\raggedright}{\thesubsubsection.}{1em}{}
\renewcommand\thesection{\arabic{section}}
\renewcommand\thesubsection{\thesection.\arabic{subsection}}
\renewcommand\thesubsubsection{\thesubsection.\arabic{subsubsection}}

\usepackage{epstopdf}
\usepackage[utf8]{inputenc}

\usepackage{hyperref}
\usepackage{xstring}
\usepackage{caption}
\usepackage{subcaption}

\usepackage{color}

\title{Evaluating Transformer Language Models on Arithmetic Operations Using Number Decomposition}

\name{Matteo Muffo, Aldo Cocco, Enrico Bertino} 

\address{Indigo.ai \\
         Via Torino 61, Milan, Italy \\
         \{matteo, aldo, e\}@indigo.ai\\}

\abstract{
In recent years, Large Language Models such as GPT-3 showed remarkable capabilities in performing NLP tasks in the zero and few shot settings. On the other hand, the experiments highlighted the difficulty of GPT-3 in carrying out tasks that require a certain degree of reasoning, such as arithmetic operations. In this paper we evaluate the ability of Transformer Language Models to perform arithmetic operations following a pipeline that, before performing computations, decomposes numbers in units, tens, and so on. We denote the models fine-tuned with this pipeline with the name \textit{Calculon} and we test them in the task of performing additions, subtractions and multiplications on the same test sets of GPT-3. Results show an increase of accuracy of 63\% in the five-digit addition task. Moreover, we demonstrate the importance of the decomposition pipeline introduced, since fine-tuning the same Language Model without decomposing numbers results in 0\% accuracy in the five-digit addition task.\\
 \\ \newline  } 

\begin{document}

\maketitleabstract


\section{Introduction} \label{sec:intro}
The publication of GPT-3 \cite{gpt3} had a relevant impact on Natural Language Processing, showing that it is possible to leverage a Large Language Model (LLM) to perform downstream tasks in the zero and few shot setting. However, although GPT-3 showed on-the-fly reasoning capabilities in tasks such as two or three-digit operations, it struggles with five-digit operations. This suggests that a LLM such as GPT-3 did not effectively learn to perform arithmetic operations and is not able to generalize its ability to perform sums or subtractions to any number of digits.

With this work we want to assess if Transformer Language Models have enough reasoning capabilities to learn to perform arithmetic operations of unseen numbers. To do so, we introduce \textit{Calculon}, a GPT-2 \cite{gpt2} model fine-tuned to perform arithmetic operations between \textit{decomposed} numbers. In particular, Calculon is trained to perform arithmetic operations following a pipeline that decomposes the numbers in digit form (e.g. 18954 = 4 units, 5 tens, 9 hundreds, 8 thousands, 1 tens of thousands). The underlying idea is to teach LMs to do computations as children learn at school, processing units with units, tens with tens, and so on. Following this pipeline, Calculon reaches remarkable levels of accuracy, even in the four and five-digit tasks where GPT-3 obtains poor results.

To validate the importance of the pipeline here proposed, we fine-tune a GPT-2 model on the same datasets of Calculon without decomposing numbers. In this setting, the fine-tuned GPT-2 network reaches very poor performances on the four and five digits tasks, demonstrating that the decomposition pipeline is a valid approach to make LMs effectively learn arithmetic. Finally, we experiment if it is possible to improve the performances of GPT-3 with the proposed decomposition pipeline via few-shot priming (no parameters update). The results obtained are worse than those of the original GPT-3 publication.

In light of the experiments carried out we conclude that Transformer Language Models have enough reasoning capabilities to effectively learn to perform addition and subtraction operations, but a higher level of reasoning is required in order to learn to perform multiplications.

In section \ref{sec:related} we review literature related to our work, in section \ref{sec:model} we describe the decomposition pipeline that we propose, in section \ref{sec:data} we describe the data used to fine-tune LMs while in section \ref{sec:results} we present and discuss the results obtained. We provide data and code used to reproduce the experiments\footnote{Available at:\\ https://github.com/mmuffo94/TransformerLM\_arithmetics}. Our code is based on the Huggingface Transformers library \cite{wolf2020huggingfaces}.

\section{Related Work} \label{sec:related}

Numeracy capabilities of NLP models have been widely experimented in literature. Numeracy benchmarks in NLP range from Arithmetic Word Problems \cite{hendrycks2021measuring} to Magnitude Comparison \cite{naik-etal-2019-exploring,wallace-etal-2019-nlp} and Measurement Estimation \cite{zhang-etal-2020-language-embeddings}. We refer to \newcite{numeracy_survey} for a complete survey about the topic. \newcite{saxton2019analysing} propose an analysis about mathematical reasoning abilities of neural NLP models by testing them on several mathematical problems such as finding the solution of equations or computing derivatives. The conclusion of this work is that a Transformer LM obtains moderate performances in the tasks analysed, suggesting that there is large room for improving mathematical reasoning capabilities of generative NLP models. \newcite{gpt3} introduced the task of performing computations by generating a numeric answer given an input prompt in natural language. In this work, the authors test the ability of GPT-3 to perform additions, subtractions, and multiplications in the zero and few-shot settings, focusing on numbers from 1 to 5 digits. Results show high levels of accuracy in two- and three-digit addition and subtraction operations, followed by low levels of accuracy as digits increase (four and five).

To complete this section, we include papers studying how different tokenization techniques affect numeracy capabilities of NLP models. \newcite{numeracy_survey} underline that sub-word tokenizers such as BPE \cite{bpe} and WordPiece \cite{wordpiece} split numbers in arbitrary tokens (e.g. \texttt{1234} can be split in \texttt{12-34}, \texttt{1-234}, \dots). Moreover, \newcite{wallace-etal-2019-nlp} demonstrate that sub-word representations provided by BERT \cite{devlin2019bert} are less effective in representing numbers with respect to character-level representations used by ELMo \cite{elmo} when probing token embedding methods in numeracy tasks. Similarly, \newcite{geva-etal-2020-injecting} show that representing numbers with a digit-level tokenizer instead of WordPiece improves the performances of their GenBERT model. Compared to these publications, with our decomposition pipeline we propose an alternative way of representing numbers for LMs. \newcite{nogueira2021investigating} propose a work similar to ours in which they fine-tune a T5 model \cite{t5}  to perform arithmetic operations, providing numbers written in different forms as input to the model. However, there is a crucial difference between \newcite{nogueira2021investigating} and our work: while the former provides manipulated numbers as input to the Transformer LM, in our work we use inputs without any manipulation and we assume that the LM will be able to generate decomposed numbers. We believe that our approach, compared to the work of \newcite{nogueira2021investigating}, will produce more robust results due to the fact that no pre-processing operation is performed over input data.

\section{Methodology} \label{sec:model}

As outlined in section \ref{sec:intro}, in this work we want to assess if Transformer Language Models have the reasoning capabilities to learn how to perform arithmetic operations between unseen numbers composed of a variable amount of digits. In particular, consistently with \newcite{gpt3}, we focus on the tasks of summing and subtracting 2, 3, 4, and 5 digit numbers and multiplying 2 digit numbers. 
In our experiments we fine-tuned a pre-trained GPT-2 Language Model\footnote{Available at \texttt{https://huggingface.co/gpt2}} \cite{gpt2} to perform computations using different approaches. We underline that our experiments are conducted in a sequence-to-sequence framework, in which the Language Model receives a string as input and provides a string as output that may contain the number corresponding to the correct answer.

The main approach we experimented is denoted as \textit{decomposition pipeline}. The idea is to teach the LM to do calculations in the same way children are taught in school. For instance, if we consider the 5 digit addition task, given the two numbers involved in the sum and the relative result, we generate the input string as follows. First, we translate both numbers in their decomposition form (e.g. \texttt{868} becomes \texttt{8 units, 6 tens, 8 hundreds}), then we sum the decomposed numbers to obtain a decomposed number as the output. Finally we reconstruct the number representing the result of the addition from its decomposed form. We underline again that all these steps are written in natural language and the string constructed will be an observation of the training set used to fine-tune GPT-2. We will refer to LMs fine-tuned with this approach under the name of \textit{Calculon}.

The second approach that we experimented, named as \textit{baseline}, is an approach in which no manipulation is done on numbers. An observation relative to this method will be simply a string containing the two numbers involved in the computation followed by the final result.

In section \ref{sec:related} we mentioned the works of \newcite{wallace-etal-2019-nlp} and \newcite{geva-etal-2020-injecting}, which evidenced that character-level tokenizers are preferable to sub-word tokenizers when processing numbers. For this reason we experiment with another approach, denoted as \textit{spaced pipeline}, in which we want to assess if a transformer LM can tokenize the digits singularly and solve the operations. An observation relative to this approach will be a string where we transform the two numbers into the \textit{spaced} form (e.g. \texttt{868} becomes \texttt{8 6 8}), then we compute the operation between the \textit{spaced} numbers and finally we reconstruct the resulting number starting from its \textit{spaced} form. The idea behind this approach is that the spacing of the digits allows the BPE tokenizer \cite{bpe} used by GPT-2 to tokenize each digit singularly.

\begin{table*}[]
    \centering
    \begin{tabular}{p{0.12\textwidth}p{0.8\textwidth}}
    \hline
         Approach & Observation \\
    \hline
         \centering Calculon & {\small \textit{\textbf{Compute with pipeline 1201 plus 1302.} Translate from number to decomposition: 1201 = 1 units, 0 tens, 2 hundreds, 1 thousands. Translate from number to decomposition: 1302 = 2 units, 0 tens, 3 hundreds, 1 thousands. Sum 1 units, 0 tens, 2 hundreds, 1 thousands + 2 units, 0 tens, 3 hundreds, 1 thousands = 3 units, 0 tens, 5 hundreds, 2 thousands. Translate from decomposition to number: 3 units, 0 tens, 5 hundreds, 2 thousands = 2503}} \\ \\
         
         \centering Baseline & {\small \textit{\textbf{Compute 1201 plus 1302.} Final result = 2503}}\\ \\
         
         \centering Spaced & {\small \textit{\textbf{Compute 1201 plus 1302.} 1 2 0 1 plus 1 3 0 2 = 2 5 0 3. Final result = 2503}} \\
    \hline
          
    \end{tabular}
    \caption{Examples of addition training observations for the considered approaches. Bold sub-strings represent input prompts provided to LMs at inference time. The same examples for the subtraction and multiplication tasks can be obtained substituting \texttt{\{plus, +, sum\}} with \texttt{\{minus, -, subtract\}} and \texttt{\{times, *, multiply\}} respectively.}
    \label{tab:decomp}
\end{table*}

At inference time, for all the approaches, the input for the fine-tuned model is a string containing two numbers and an arithmetic operation. If at the end of the generated string there is the number corresponding to the result of the operation, the observation is considered correct.
In table \ref{tab:decomp} we provide examples of training observations and inference inputs for each of the studied approaches.

The last experiment that we conducted is about GPT-3 Language Model. In particular, with this set of tests we want to assess if GPT-3 can benefit from the decomposition pipeline in a few-shot setting (without any parameter update). In this case the experiments consist of evaluating GPT-3 in the same tasks mentioned at the beginning of this section, but providing in the input prompt\footnote{A full GPT-3 input prompt reported in  Appendix A} only 4 few-shot examples with the \textit{decomposition pipeline} that we introduced.

\section{Data and training details} \label{sec:data}
For the addition and subtraction operations, we generate training sets of 12000 observations each. In particular for each $N\in\{3,4,5\}$ we randomly sample 3000 couples of integer numbers $(n_1, n_2)_i$, with  $(n_1, n_2)_i\in\{10^{N-1},\dots,10^{N}-1\}^2, \forall i\in\{1,\dots,3000\}$. Similarly, for $N=2$ we randomly sample 3000 couples of numbers $(n_1, n_2)_i \in \{0,\dots,99\}^2$ (one-digit numbers are included). We then join all the couples created (obtaining a set of 12000 couples) and we compute the results of the operations. At the end of this step, we obtain two vectors of results, $\mathbf{r_{+}}$ and $\mathbf{r_{-}}$, where $r_{+,i}=n_{1,i}+n_{2,i}$ and $r_{-,i}=n_{1,i}-n_{2,i}, \forall i\in\{1,\dots,12000\}$. Finally, given a triplet $(n_1,n_2,r)_i$, we generate a string according to the procedures described in section \ref{sec:model}, depending on the selected approach. For the multiplication, we generate training sets following the same procedure explained above but, instead of sampling 12000 couples, we sample 3000 couples of numbers from the set $\{0,\dots,99\}^2$ because we will only test the multiplications between two-digit numbers.

At the end of this procedure we obtain 9 training sets, each of which corresponding to a combination operation-approach (e.g. addition-decomposition), that we use to fine-tune as many Language Models. Now, we want to underline some points relative to the generated training sets. First, by fixing the operation and varying the approach, the same couples of numbers are used to generate strings, so that couples of numbers are sampled once for each operation. Second, none of the couples present in a training set is in the test set relative to the same operation. The test sets used to evaluate our fine-tuned Language Models are the same used to evaluate GPT-3 in the arithmetic tasks\footnote{Test sets publicly available at \texttt{https://github.com/openai/gpt-3}} \cite{gpt3}.

The GPT-2 models fine-tuned in our experiments are \textit{GPT-2 Small} architectures, which count $\sim$117M parameters. The GPT-3 model evaluated in our experiments corresponds to the biggest architecture proposed in \newcite{gpt3}, which counts $\sim$175B parameters.
We fine-tune each GPT-2 Language Model for 25 epochs with an initial learning rate of $10^{-4}$ and a batch size of 32, using Adam \cite{adam} as optimizer. For the experiments on GPT-3, due to limited resources available on the dedicated API, we evaluate the model only on the first 100 observations of each test set.
We adopt a greedy decoding strategy for the GPT-2 models and a random sampling strategy with temperature=0.7 for the GPT-3 generations.

\section{Results and discussion} \label{sec:results}
In table \ref{tab:results} we show the results obtained with the experiments described in section \ref{sec:model}.

\begin{table*}[]
    \centering
    \begin{tabular}{c c c c c c c c c c}
    \hline
         Approach & 2D+ & 3D+ & 4D+ & 5D+ & 2D- & 3D- & 4D- & 5D- & 2Dx \\
    \hline
         Calculon & 99.75 & \textbf{81.95} & \textbf{80.05} & \textbf{72.85} & \textbf{100.00} & 81.35 & \textbf{78.60} & \textbf{75.65} & 14.65 \\
         Baseline & 53.35 & 5.60 & 0.05 & 0.00 & 22.30 & 1.60 & 0.05 & 0.00 & 5.25 \\
         Spaced & 90.10 & 77.75 & 67.10 & 57.95 & 45.20 & 11.80 & 1.35 & 0.15 & 5.1 \\
         GPT-3 FS &  \textbf{100.0} & 80.4 & 25.5 & 9.3 & 98.9 & \textbf{94.2} & 26.8 & 9.9 & \textbf{29.2} \\
         GPT-3 FS decomp & 3.0 & 0.0 & 0.0 & 0.0 & 3.0 & 0.0 & 0.0 & 0.0 & 1.0 \\
    \hline
          
    \end{tabular}
    \caption{Accuracy scores obtained in the experiments described in section \ref{sec:model}. \textit{\{2, 3, 4, 5\}D\{+, -\}} represents 2, 3, 4 or 5 addition or subtraction tasks. \textit{2Dx} represents the 2 digit multiplication task. \textit{GPT-3 FS} refers to results obtained by GPT-3 in the few-shot setting (results in this row are those obtained in Brown et al., 2020). \textit{GPT-3 FS decomp} refers to results obtained by GPT-3 using the decomposition pipeline in few shot examples. Results relative to this last experiment are obtained over the first 100 observations of each test set exclusively.}
    \label{tab:results}
\end{table*}

The GPT-2 model fine-tuned without decomposition (\textit{Baseline} row) obtains low accuracy scores in all tasks except two-digit addition, where achieves 53.35 accuracy. In particular, in the 4 and 5 addition and subtraction tasks it achieves zero or near-zero accuracy. This demonstrates that, without decomposing numbers, a GPT-2 Language Model is not able to learn to perform computations, especially between numbers with a higher number of digits. On the other hand, Calculon obtains high accuracy scores in all the tasks tested with the exception of 2Dx. This demonstrates that fine-tuning using the proposed decomposition pipeline effectively makes possible for a transformer Language Model to learn to do calculations. Here, we underline again that none of the couples of numbers composing the training set are in the relative test set, and hence we can conclude that Calculon has effectively learned to sum units with units, tens with tens, and so on and manage to perform arithmetic operations between unseen numbers. However, the results in the two digit multiplication task are poor, suggesting that number decomposition is not sufficient to solve this task and probably higher reasoning capabilities are needed to multiply numbers.

\begin{figure*}[t] 
     \centering
     \begin{subfigure}[b]{1\textwidth}
         \centering
         \includegraphics[width=\textwidth]{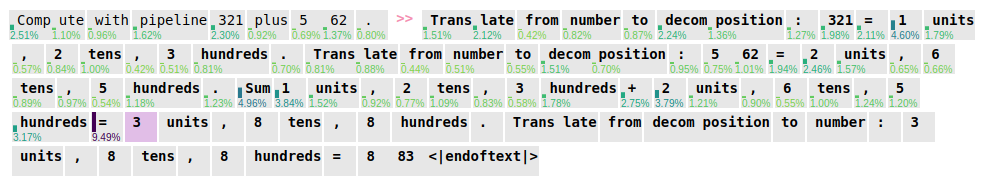}
         \caption{Input saliency scores of tokens preceding the digit \textit{3} (in purple).}
         \label{fig:saliency_a}
     \end{subfigure}
     \hfill
     \begin{subfigure}[b]{1\textwidth}
         \centering
         \includegraphics[width=\textwidth]{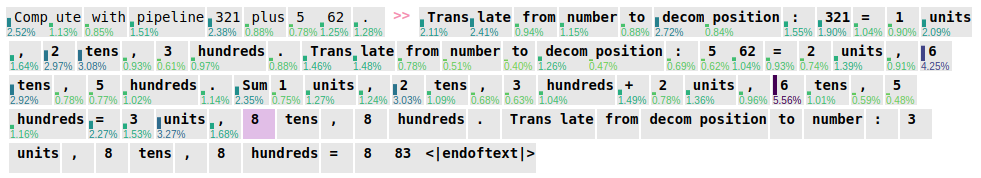}
         \caption{Input saliency scores of tokens preceding the digit \textit{8} (in purple).}
         \label{fig:saliency_b}
     \end{subfigure}
     \hfill
     \begin{subfigure}[b]{1\textwidth}
         \centering
         \includegraphics[width=\textwidth]{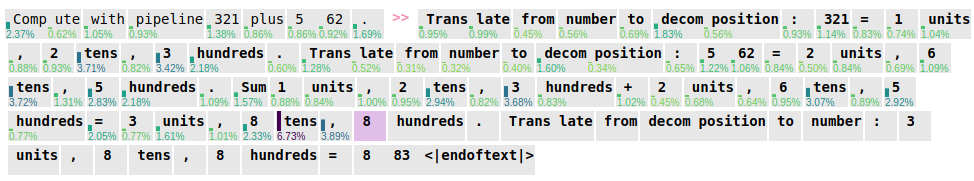}
         \caption{Input saliency scores of tokens preceding the digit \textit{8} (in purple).}
         \label{fig:saliency_c}
     \end{subfigure}
        \caption{Input saliency scores for the \textit{addition-Calculon} model.}
    \label{fig:saliency}
\end{figure*}

In figure \ref{fig:saliency} we report input saliency scores obtained from the \textit{addition-Calculon} model. These scores are obtained using the gradient-based method described in \newcite{saliency} and can be interpreted as the influence of each token related to the generated one (highlighted in purple in pictures). Figures and saliency scores are obtained using the Ecco package \cite{ecco}. We notice that, when analysing the unit digit resulting from the addition (figure \ref{fig:saliency_a}) the digits with highest saliency scores are those relative to units in the two numbers which are summed (namely \textit{1} and \textit{2}). A similar behavior is present in figures \ref{fig:saliency_b} and \ref{fig:saliency_c} for tens and hundreds respectively. This indicates that Calculon models effectively learn that units must be summed with units and so on in order to perform a correct addition.

We wanted to assess that the benefits brought by the decomposition pipeline are not due only to the fact that digits are tokenized separately. We then conducted the \textit{Spaced} experiments, in which we separated digits with a blank space but without indicating if a digit corresponds to units, tens, and so on. The results obtained with these experiments (\textit{Spaced} row) show a remarkable improve of accuracy with respect to the baseline in the addition tasks followed by a small improve in the subtraction tasks. However Calculon maintains a further improvement with respect to the Spaced approach, with an accuracy gain of almost 15\% in the 5D+ task and 75\% in the 5D- task. This indicates that providing magnitude information of digits when decomposing numbers significantly helps LMs when performing arithmetic operations. Moreover, the gap in accuracy scores between baseline and Spaced approaches is consistent with the work of \newcite{wallace-etal-2019-nlp} and \newcite{geva-etal-2020-injecting}. Lastly, observing the results obtained by GPT-3, we notice that decomposing numbers in the few-shot examples (row \textit{GPT-3 FS decomp}) leads to much lower results with respect to the original GPT-3 results (row \textit{GPT-3 FS}), in which no manipulation is performed over numbers in the few-shot examples. We hypothesize that receiving a quite long input prompt GPT-3 loses the focus on the task of performing an arithmetic operation and the decomposition prevents it from leveraging on the calculations seen during the pre-train.

\section{Conclusions and future work}
In this work we presented Calculon, a GPT-2 Language Model fine-tuned to perform arithmetic operations following a pipeline that decomposes numbers before the computations. We showed that a Transformer Language Model can effectively learn to perform calculations generalizing to unseen numbers when fine-tuned with the decomposition pipeline proposed. Moreover, when we fine-tune on the same task but without the number decomposition, the same GPT-2 network reaches very poor results, proving that the decomposition pipeline effectively brings benefit during the training. On the other hand, we showed that adopting the same decomposition pipeline when providing few shot examples to GPT-3 leads to very bad results, suggesting that decomposition does not bring the same benefit in the few shot setting. We demonstrated that, by decomposing numbers, a Transformer LM such as GPT-2 has the reasoning capabilities to learn during fine-tuning the rules and procedures to perform additions and subtractions, but the same does not hold for the multiplication operation. 

There may be a variety of future works related to our experiments. First, consistently with \newcite{gpt3}, we tested LMs on up to five-digit operations, but it can be interesting to assess if decomposition brings the same benefit for operations involving a higher number of digits. Secondly, it may be interesting to evaluate why all the tested models struggle with multiplication and if the latter is an operation that requires a level of reasoning unattainable by current linguistic models. In addition, it will be useful to investigate how the number of observations in the training sets affects the performances of fine-tuned GPT-2 models in the studied tasks. Regarding the experiments on GPT-3, it can be investigated whether a more compact formulation of the decomposition pipeline can also bring benefits in the context of few-shot learning. Finally, it can be interesting to extend the experiments presented in this work to other Transformer LMs such as T5 \cite{t5} and BART \cite{bart}.

\newpage
\section*{Appendix: GPT-3 input prompt} \label{sec:appendix_a}
In the following lines we report the addition input prompt used when evaluating GPT-3 with decomposition (row \textit{GPT-3 FS decomp} of table 2).\\

\texttt{This application makes an arithmetic operation decomposing the input numbers.\\
\#\#\#\\
Compute with pipeline 28 plus 39. 
Translate from number to decomposition: 28 = 2 tens, 8 units. 
Translate from number to decomposition: 39 = 3 tens, 9 units. 
Sum 8 units, 2 tens + 9 units, 3 tens = 7 units, 6 tens. 
Translate from decomposition to number: 6 tens, 7 units = 67\\
\#\#\#\\
Compute with pipeline 804 plus 121. 
Translate from number to decomposition: 804 = 8 hundreds, 0 tens, 4 units. 
Translate from number to decomposition: 121 = 1 hundreds, 2 tens, 1 units. 
Sum 4 units, 0 tens, 8 hundreds + 1 units, 2 tens, 1 hundreds = 5 units, 2 tens, 9 hundreds. 
Translate from decomposition to number: 9 hundreds, 2 tens, 5 units = 925\\
\#\#\#\\
Compute with pipeline 1201 plus 1302. 
Translate from number to decomposition: 1201 = 1 thousands, 2 hundreds, 0 tens, 1 units. 
Translate from number to decomposition: 1302 = 1 thousands, 3 hundreds, 0 tens, 2 units. 
Sum 1 units, 0 tens, 2 hundreds, 1 thousands + 2 units, 0 tens, 3 hundreds, 1 thousands = 3 units, 0 tens, 5 hundreds, 2 thousands. 
Translate from decomposition to number: 2 thousands, 5 hundreds, 0 tens, 3 units = 2503\\
\#\#\#\\
Compute with pipeline 97734 plus 86328. 
Translate from number to decomposition: 97734 = 9 tens of thousands, 7 thousands, 7 hundreds, 3 tens, 4 units. 
Translate from number to decomposition: 86328 = 8 tens of thousands, 6 thousands, 3 hundreds, 2 tens, 8 units. 
Sum 4 units, 3 tens, 7 hundreds, 7 thousands, 9 tens of thousands + 8 units, 2 tens, 3 hundreds, 6 thousands, 8 tens of thousands = 2 units, 6 tens, 0 hundreds, 4 thousands, 8 tens of thousands, 1 hundreds of thousands. 
Translate from decomposition to number: 1 hundreds of thousands, 8 tens of thousands, 4 thousands, 0 hundreds, 6 tens, 2 units = 184062\\
\#\#\#\\
Compute with pipeline \{number1\} plus \{number2\}.}\\

\texttt{\{number1\}} and \texttt{\{number2\}} are replaced by the numbers involved in the computation. 
The string is composed by 4 few-shot examples which follow the decomposition pipeline introduced in section 3. 

\section{Bibliographical References}\label{reference}

\bibliographystyle{lrec2022-bib}
\bibliography{lrec2022-example}


\end{document}